\documentclass[pdflatex,sn-mathphys-num]{sn-jnl}% Math and Physical Sciences Numbered Reference Style
\usepackage{graphicx}%
\usepackage{multirow}%
\usepackage{amsmath,amssymb,amsfonts}%
\usepackage{amsthm}%
\usepackage{mathrsfs}%
\usepackage[title]{appendix}%
\usepackage{xcolor}%
\usepackage{textcomp}%
\usepackage{manyfoot}%
\usepackage{booktabs}%
\usepackage{algorithm}%
\usepackage{algorithmicx}%
\usepackage{algpseudocode}%
\usepackage{listings}%
\theoremstyle{thmstyleone}%
\theoremstyle{thmstyletwo}%

\theoremstyle{thmstylethree}%

\begin{document}

\title[Weakly Supervised Tuberculosis Localization in Chest X-rays through Knowledge Distillation]{Weakly Supervised Tuberculosis Localization in Chest X-rays through Knowledge Distillation}


\author*{\fnm{Marshal~Ashif} \sur{Shawkat}}\email{marshalashif21@gmail.com}
\equalcont{These authors contributed equally to this work.}

\author{\fnm{Moidul} \sur{Hasan}}\email{moidulhasan1821@gmail.com}
\equalcont{These authors contributed equally to this work.}

\author{\fnm{Taufiq} \sur{Hasan}}\email{taufiq@bme.buet.ac.bd}

\affil*{\orgdiv{Department of Biomedical Engineering}, \orgname{Bangladesh University of Engineering and Technology}, \orgaddress{\street{ECE Building, West Palashi}, \city{Dhaka}, \postcode{1205},  \country{Bangladesh}}}


\abstract{Tuberculosis (TB) remains one of the leading causes of mortality worldwide, particularly in resource-limited countries. Chest X-ray (CXR) imaging serves as an accessible and cost-effective diagnostic tool but requires expert interpretation, which is often unavailable. Although machine learning models have shown high performance in TB classification, they often depend on spurious correlations and fail to generalize. Besides, building large datasets featuring high-quality annotations for medical images demands substantial resources and input from domain specialists, and typically involves several annotators reaching agreement, which results in enormous financial and logistical expenses. This study repurposes knowledge distillation technique to train CNN models reducing spurious correlations and localize TB-related abnormalities without requiring bounding-box annotations. By leveraging a teacher-student framework with ResNet50 architecture, the proposed method trained on TBX11k dataset achieve impressive 0.2428 mIOU score. Experimental results further reveal that the student model consistently outperforms the teacher, underscoring improved robustness and potential for broader clinical deployment in diverse settings.}




\maketitle

\section{Introduction}\label{sec1}

Tuberculosis (TB) is a highly contagious infectious disease that has consistently remained one of the leading causes of morbidity and mortality worldwide. TB continues to affect an estimated 10 million people annually, resulting in approximately 1.4 million deaths each year\cite{WHO2020}. The disease primarily targets the lungs and is transmitted through airborne particles, making it highly communicable in densely populated and poorly ventilated environments. Early detection is therefore critical not only for improving patient outcomes but also for limiting the spread of the disease within communities. However, accurate and timely diagnosis of TB remains a persistent challenge, particularly in low-resource settings\cite{Bekmurzayeva2013}. The current gold standard for TB diagnosis involves microscopic examination of sputum samples and culturing of Mycobacterium tuberculosis. This process requires biosafety level-3 (BSL-3) laboratory infrastructure due to the infectious nature of the bacterium. Unfortunately, such facilities are often unavailable in many hospitals across developing nations, creating significant diagnostic bottlenecks and delaying treatment initiation.\par

In contrast, chest X-ray (CXR) imaging offers a faster and more accessible screening alternative. It is the most commonly used imaging modality in TB screening and is particularly effective in detecting pulmonary abnormalities indicative of TB infection. The World Health Organization recommends CXR as a preliminary screening tool, especially in mass-screening or resource-limited scenarios\cite{WHO_ChestRadiography}. When interpreted correctly, CXRs can facilitate early diagnosis, timely intervention, and reduced transmission. However, accurate interpretation of CXR images is highly dependent on expert radiologists, and even seasoned professionals may struggle to detect subtle manifestations of TB due to inter-observer variability and the inherent limitations of human visual perception.\cite{Satia2013} \par

Machine learning has emerged as a promising solution in the field of medical image analysis. These advances have naturally extended into the domain of tuberculosis detection using Chest X-ray images, where machine learning models like Convolutional Neural Network (CNN) have demonstrated high diagnostic accuracy leveraging its superior feature extraction and representation learning capabilities.\cite{Huy2023} \par

However, despite their impressive performance, many deep learning models suffer from poor generalization when applied to unseen datasets. This is often attributed to the models learning spurious correlations or "shortcuts"—such as the presence of hospital-specific tokens, scanner-specific artifacts, or patient demographics—that are not causally related to the disease itself \cite{Geirhos2020}. These shortcuts can inflate performance on training and internal validation sets but lead to model failures in real-world deployment, especially across different geographic or clinical domains. As such, there is a growing need for robust, interpretable, and generalizable models that not only excel in controlled experiments but also maintain high reliability in diverse and unseen clinical environments. Moreover, constructing large datasets with high-quality medical image annotations is particularly challenging, resource-intensive, requires domain experts, and often needs multiple annotators to achieve consensus, incurring massive financial and logistical costs \cite{Wang2021}.\par

\begin{figure}[htbp]
\centering
\includegraphics[width=0.9\linewidth]{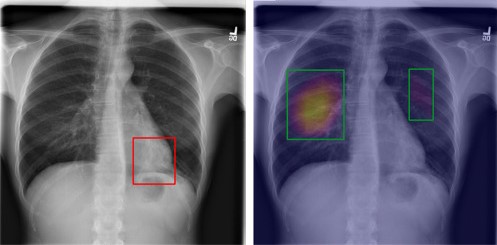}

\caption{Deep learning models often prioritize spurious correlations over disease-specific features. Grad-CAM highlights an incorrect chest region \cite{Luo2022}.}
\label{fig:nontargeted}
\end{figure}

\begin{figure}[htbp]
\centering

\begin{minipage}{0.45\linewidth}
\centering
\includegraphics[width=\linewidth]{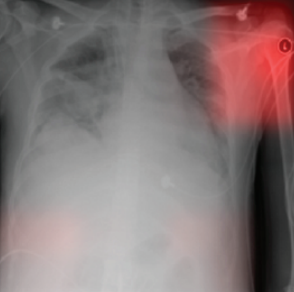}
\label{fig:wrong_focus}
\end{minipage}
\hfill
\begin{minipage}{0.45\linewidth}
\centering
\includegraphics[width=\linewidth]{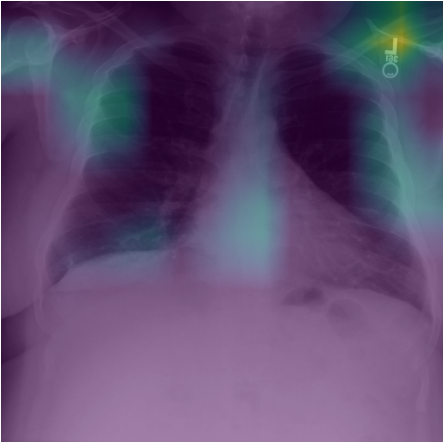}
\label{fig:tag_bones}
\end{minipage}

\caption{Deep learning models may focus on hospital tokens rather than the lung region \cite{Geirhos2020, Jabbour2020}.}
\label{fig:wrongfocus_combined}
\end{figure}

This study employs knowledge distillation (KD), a powerful model compression and transfer learning technique, to develop an effective tuberculosis (TB) detection model that can accurately localize lesions in chest X-ray (CXR) images without the need for bounding-box annotations. Knowledge distillation works by training a smaller, simpler model (called the student) to mimic the behavior of a larger, more complex, and accurate model (called the teacher)\cite{Hinton2015}. The teacher model is first trained on a given task and produces output predictions, such as soft class probabilities that capture richer information than standard one-hot labels. The student model then learns from these soft targets.\par

Our main contribution is two fold. Firstly, we demonstrate that knowledge distillation can be effectively used to teach CNN-based models to focus on TB-specific features, reducing spurious correlations. Secondly, we localize tuberculosis in chest X-rays without bounding box information. This second novelty is particularly significant, as it eliminates the dependency on expensive bounding box annotations, which typically require multiple expert radiologists to delineate lesion boundaries precisely. By generating bounding boxes solely from class labels and Grad-CAM heatmaps, our method drastically lowers the barriers to dataset creation, enabling scalable TB screening in resource-constrained environments. We hope that this research will shed new light on the power of knowledge distillation apart from model compression.

\section{Related Work}\label{sec2}
Alshmarani et al. proposed a deep learning architecture using a pre-trained VGG19 model followed by three blocks of Convolutional Neural Network (CNN) for feature extraction, and a fully connected network for multi-class classification of lung diseases. The proposed model achieved superior performance with 96.48\% accuracy \cite{Alshmrani2023}. The DNN model\cite{Liu2023} uses a DenseNet121 architecture as its backbone, which was pre-trained on large public datasets (MIMIC-CXR\cite{Johnson2019} and CheXpert\cite{Irvin2019}) and then fine-tuned on an in-house dataset of 1500 CXRs from TB, NTM-LD, and Imitator patients. The DNN model achieved a higher classification accuracy (66.5\% ± 2.5\%) compared to both senior (50.8\% ± 3.0\%) and junior pulmonologists (47.5\% ± 2.8\%) on the internal test set. Vanitha et al. present a Vision Transformer model\cite{Dosovitskiy2021}  with a Conv2D stem and Grad-CAM interpretability that achieves state-of-the-art accuracy, recall, and F1-scores for tuberculosis detection from chest X-rays \cite{Vanitha2025}. El-Ghany et al. propose a ViT with Principal Component Analysis (PCA) hybrid system for TB classification, yielding very high metrics (accuracy 99.84\%) \cite{ElGhany2024}. These models achieve impressive accuracy but do not consider the detrimental effect of spurious correlations learned by the CNN models.\par

Geirhos et al. showed that a deep learning model failed to recognize pneumonia from  X-ray scans of a new hospital because the model learned to predict pneumonia based on hospital-specific attributes\cite{Geirhos2020}. Using state-of-the-art techniques in explainable AI, DeGrave et al. demonstrate that recent deep learning systems to detect COVID-19 from chest radiographs rely on confounding factors rather than medical pathology, creating an alarming situation in which the systems appear accurate, but fail when tested in a new hospital\cite{DeGrave2021}. Luo et al. illustrate that a DL model trained with radiograph-level annotations is prone to shortcut learning (ie, using unintended patterns for decision-making\cite{Luo2022}. It can be observed that CheXNet’s Grad-CAM might not precisely cover the intended lesions and sometimes even attend to false-positive regions (Fig \ref{fig:nontargeted}). Studying networks trained with pneumonia datasets, Catalá et al. observed heatmaps often highlighted areas of the image which did not contain lung tissue, rather model learn from simple features like white rectangular box at the top corner \cite{Catala2021}. Jabbour et al. reveal that deep learning models trained on chest X-rays can exploit spurious correlations such as patient demographics, treatments, or preprocessing artifacts, leading to shortcut learning and poor generalization across hospitals\cite{Jabbour2020} (Fig \ref{fig:tag_bones}).\par

Wang et al. introduces a weakly supervised chest X-ray abnormality localization framework using VMamba\cite{Liu2024VMamba} with non-linear modulation, FPM fusion, and a novel foreground control loss\cite{Wang2024}. Termritthikun uses knowledge distillation (KD) method to create a compact deep learning model for real-time multi-label chest X-ray classification\cite{Termritthikun2024}. Akhter et al. tackles the challenge of diagnosing low-resolution chest X-rays in resource-limited settings by transferring knowledge from a high-resolution teacher model to a low-resolution student model\cite{Akhter2024}. Kabir et al. develops a knowledge distillation framework designed to improve the detection of respiratory diseases from chest X-ray images\cite{Kabir2024}. Park et al. proposes a self-evolving vision transformer framework that combines self-supervision and self-training via knowledge distillation to leverage unlabeled chest X-rays\cite{Park2022}. While prior works have applied knowledge distillation for model compression, efficiency in low-resolution settings, or multi-label classification, none have explicitly utilized it to mitigate spurious correlations in TB-specific CXR representations.

\section{Methodology}\label{sec3}

\begin{figure}[t]
\centering
\includegraphics[width=0.9\linewidth]{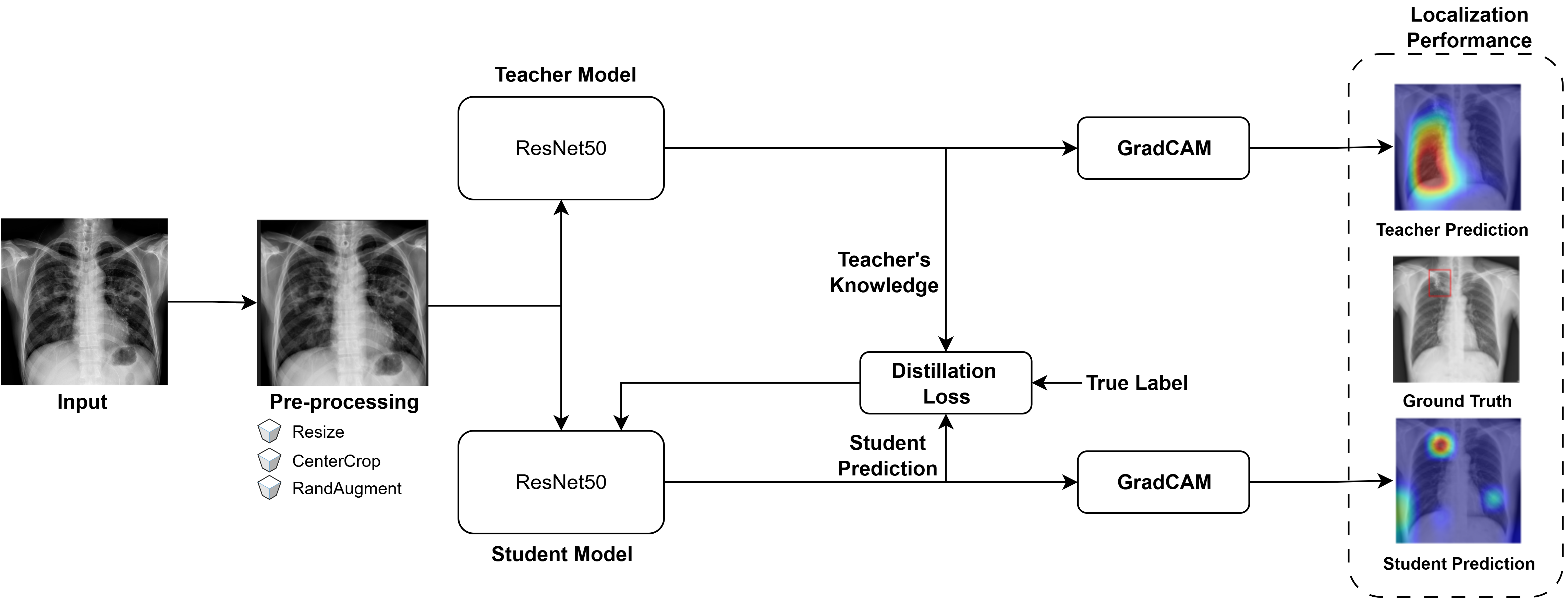}
\caption{A graphical overview of the proposed framework. Following a series of preprocessing steps, the training images are input into both the teacher and student networks, each based on a pretrained ResNet50 architecture. The student network is optimized using a distillation loss function that combines a weighted soft-target loss with cross-entropy loss. Grad-CAM visualization demonstrates enhanced localization performance under this framework.}
\label{fig:overview}
\end{figure}

\subsection{Dataset}\label{subsec1}
The TBX11k dataset \cite{Liu2024} contains 8,400 publicly available chest X-ray images. Among them, 800 CXRs are TB images. The rest of the images are from healthy individuals or individuals suffering from chest diseases other than tuberculosis. The TB CXRs contain bounding-box information on the TB locations in the chest regions.
\subsection{Implementation Details}\label{subsec2}
The TBX11k dataset already specifies which images belong to the training set and the validation set. The ground truth labels of the test set is not publicly available. Thus we treat the validation set as the test set. There are a total of 6600 CXR images in the training set. Among them 600 are TB images and the rest 6000 are treated as non-TB images. The test (validation) set contains 200 TB images and 1600 Non-TB images making a total of 1800 CXR images.\\

% \begin{figure}
%     \centering
%     \includegraphics[width=0.9\linewidth]{images/dataset_split.png}
%     \caption{Distribution of tuberculosis (TB) and non-tuberculosis (Non-TB) chest X-ray (CXR) images in the TBX11K dataset. The dataset provides predefined training and validation splits, where the validation set is used as the test set in this study due to the unavailability of ground truth labels for the official test set. The training set contains 6600 CXR images (600 TB and 6000 Non-TB), while the test set comprises 1800 CXR images (200 TB and 1600 Non-TB).}
%     \label{fig:dataset_split}
% \end{figure}

\subsubsection{Preprocessing}\label{subsubsec1}
The images have size of (512, 512, 3) in order of (height, width, no of channels). First, we resize the images to (256, 256, 3). Then we crop the center of the images and make them (224, 224, 3). Then we apply RandAugment (num ops = 2, magnitude = 9) as it has been shown to improve the model accuracy \cite{Cubuk2019}. Finally, we normalize the images by ImageNet mean [0.485, 0.456, 0.406] and standard deviation [0.229, 0.224, 0.225].\\
\subsubsection{Training}\label{subsubsec2}
ResNet50 \cite{He2015} architecture is chosen to train both the teacher network and the student network. We initialize our model using pretrained ImageNet model from the torchvision library. Fine-tuning pretrained models converge faster than training from scratch\cite{Matsoukas2022}. We change the last layer of the model to account for two classes (TB and No-TB). We use SEED 42 for weight initialization of the last layer. Categorical CrossEntropy loss is used to classify chest X-rays. The batch size is 64. We train both models for a maximum number of 100 epochs. Early stopping is used if the loss does not decrease for 10 consecutive epochs. The learning rate is 1e-3. The weight decay is 1e-4. We use Adam optimizer\cite{Kingma2017} with default parameters.\par
For student network, the temperature value is set to 2. The total loss for the student model is defined as follows:
\begin{equation}
\text{total loss} = (1 - \alpha) \cdot \mathcal{L}_{ST} +\alpha \cdot \mathcal{L}_{CE}
\label{eq:total_loss}
\end{equation}
where $\mathcal{L}_{CE}$ is standard cross-entropy loss and
\begin{equation}
\mathcal{L}_{ST} = {\text{soft target loss}} 
=  T^2 \, \mathcal{KL}\!\left(
    \mathrm{softmax}\!\left(\frac{\mathbf{z}_t}{T}\right)
    \,\Big\|\, 
    \mathrm{softmax}\!\left(\frac{\mathbf{z}_s}{T}\right)
\right)
\label{eq:soft_target_loss}
\end{equation}
where\\
$T$: temperature (hyperparameter). \\
$\mathcal{KL}$: Kullback–Leibler divergence. \\
$\mathbf{z}_t$: teacher logits \\
$\mathbf{z}_s$: student logits\\

We choose the value of the hyperparameter $\alpha$ 0.75.

\subsection{Bounding Box Generation}\label{subsec3}
\begin{figure}[htbp]
\centering
\includegraphics[width=0.9\linewidth]{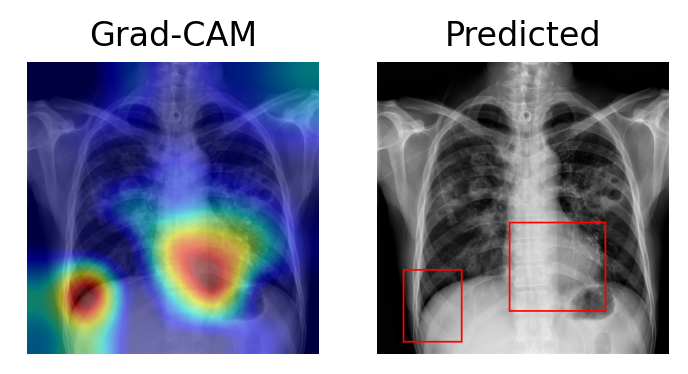}
\caption{Example of bounding box generation from Grad-CAM heatmaps. The left image shows the Grad-CAM visualization highlighting TB-suspected regions, while the right image shows the bounding boxes generated by Algorithm~\ref{alg1:gradcam_bbox}. This post-hoc bounding box extraction enables weakly supervised localization without explicit bounding box annotations during training.}
\label{fig:bbox_from_heatmap}
\end{figure}

We use the Captum library\cite{Kokhlikyan2020} to generate the Grad-CAMs\cite{Selvaraju2019}. We devise Algorithm 1 to generate bounding boxes from Grad-CAMs. The result of Algorithm 1 is shown in Figure \ref{fig:bbox_from_heatmap}. We calculate the area of the bounding box from the TBX11k training set images. Then we set the minimum area to ensure that our generated bounding boxes are not too small.
\begin{algorithm}
\caption{Bounding box generation from Grad-CAM}\label{alg1:gradcam_bbox}
\begin{algorithmic}[1]
\State Generate heatmap using Grad-CAM method
\State Normalize the heatmap
\State Create a binary mask using a threshold value
\State Identify contours
\For{each contour in contours}
    \State Compute the coordinates
    \State Compute the area
    \If{area $>$ minimum area}
        \State Include the contour in the list
    \EndIf
\EndFor
\end{algorithmic}
\end{algorithm}

This bounding box generation process represents a core novelty of our work, enabling weakly supervised localization without any explicit bounding box annotations during training. Traditional localization methods demand pixel-level or bounding box labels, which are costly to obtain in medical domains due to the need for expert consensus. By deriving bounding boxes post-hoc from Grad-CAM heatmaps produced by the distilled student model, we leverage the model's learned attention to infer lesion boundaries accurately. This approach not only bypasses the annotation bottleneck but also scales to large datasets, potentially reducing costs by orders of magnitude in resource-limited settings. The threshold and minimum area filtering ensure that only salient, TB-relevant regions are boxed, further emphasizing disease-specific focus over spurious elements.

\subsection{Evaluation Metric for Localization}\label{subsec4}
We define mean intersection over union (mIOU) according to Algorithm 2 and 3. This custom mIOU metric allows us to quantitatively assess the alignment between generated and ground truth bounding boxes, highlighting the efficacy of our annotation-free localization. By focusing on the top overlaps, it accounts for multiple lesions per image, providing a robust measure of localization accuracy in weakly supervised scenarios.

\begin{algorithm}
\caption{Mean IOU (mIOU) Score Calculation}\label{alg2_mIOU}
\begin{algorithmic}[1]
\For{each image}
    \State $\text{IOU\_LIST} \leftarrow \textsc{CalculateIoUList}(\text{GT\_Boxes}, \text{Pred\_Boxes})$
    \State Sort IOU\_LIST in descending order
    \State Keep only the top $\mathrm{len}(\text{GT\_Boxes})$ entries
\EndFor
\State Take average over all images
\end{algorithmic}
\end{algorithm}

\begin{algorithm}
\caption{\textsc{CalculateIoUList}(\text{GT\_Boxes}, \text{Pred\_Boxes})}\label{alg_calcioulist}
\begin{algorithmic}[1]
\State $\text{IOU\_LIST} \leftarrow [\,]$
\For{each $\text{GT\_ITEM} \in \text{GT\_Boxes}$}
    \For{each $\text{PRED\_ITEM} \in \text{Pred\_Boxes}$}
        \State $\text{IOU} \leftarrow \textsc{ComputeIoU}(\text{PRED\_ITEM}, \text{GT\_ITEM})$
        \State Append(IOU\_LIST, IOU)
    \EndFor
\EndFor
\Return IOU\_LIST
\end{algorithmic}
\end{algorithm}

\subsection{Evaluation Metric for Classification}\label{subsec7}
For classification, the goal is to classify each chest X-ray image into one of the two categories: TB and non-TB. To assess the classification result we use the following evaluation metrics -

     \textbf{Accuracy: }Accuracy measures the percentage of Chest X-ray images that are correctly classified.
        \begin{equation}
            \text{Accuracy} = \frac{\text{TP} + \text{TN}}{\text{TP} + \text{TN} + \text{FP} + \text{FN}}
            \label{eq:accuracy}
        \end{equation}
        Where - \\
        TP = True Positive \\
        TN = True Negative \\
        FP = False Positive \\
        FN = False Negative \\

    \textbf{AUC: }The AUC measures the area beneath the Receiver Operating Characteristic (ROC) curve which depicts the relationship between the true positive rate and the false positive rate for the tuberculosis (TB) class.\\
    
    \textbf{Sensitivity: }Sensitivity measures the proportion of tuberculosis (TB) cases that are correctly detected as TB.
        \begin{equation}
            \text{Sensitivity}  = \frac{\text{TP}}{\text{TP} + \text{FN}}
            \label{eq:sensitivity}
        \end{equation}
    
     \textbf{Specificity: }Specificity determines the percentage of non-TB cases that are correctly identified as non-TB.
        \begin{equation}
            \text{Specificity}  = \frac{\text{TN}}{\text{TN} + \text{FP}}
            \label{eq:specificity}
        \end{equation}

        \textbf{Average precision: }Average precision (AP) measures the precision for each class and then computes the mean across all classes.
        \begin{equation}
            \text{Precision} = \frac{\text{TP}}{\text{TP} + \text{FP}}
            \label{eq:precision}
        \end{equation}

     \textbf{Average recall: }Average recall (AR) measures the recall for each class and then computes the average across all classes.
        \begin{equation}
            \text{Recall} = \frac{\text{TP}}{\text{TP} + \text{FN}}
            \label{eq:recall}
        \end{equation}

Each metric is used measured in 0 to 1 scale where the closer the metric value to 1 the better.

\subsection{Hessian Matrix}\label{subsec6}
Hessian-based analysis is one of the most mathematically grounded ways to study the loss landscape of a neural network. For a neural network with parameters $\boldsymbol{\theta}$, and a loss function $L(\boldsymbol{\theta})$, the Hessian matrix is the matrix of all second-order partial derivatives:
\begin{equation}
    \mathbf{H} = \nabla_{\boldsymbol{\theta}}^2 L(\boldsymbol{\theta})
    \label{eq:hessian1}
\end{equation}

Each element is given by:
\begin{equation}
    H_{ij} = \frac{\partial^2 L}{\partial \theta_i \, \partial \theta_j}
    \label{eq:hessian2}
\end{equation}
Thus, $\mathbf{H}$ captures the local curvature of the loss surface near a point $\boldsymbol{\theta}$. \\
The trace is the sum of the diagonal elements:
\begin{equation}
    \mathrm{Tr}(\mathbf{H}) = \sum_{i=1}^{n} H_{ii} = \sum_{i=1}^{n} \lambda_i
    \label{eq:trace_hessian}
\end{equation}
Here, $\lambda_i$ is $i^{th}$ eigenvalue of the hessian matrix $\mathbf{H}$.

\section{Experimental Results and Analysis}\label{sec4}

\begin{figure}[htbp]
\centering
\includegraphics[width=0.9\linewidth]{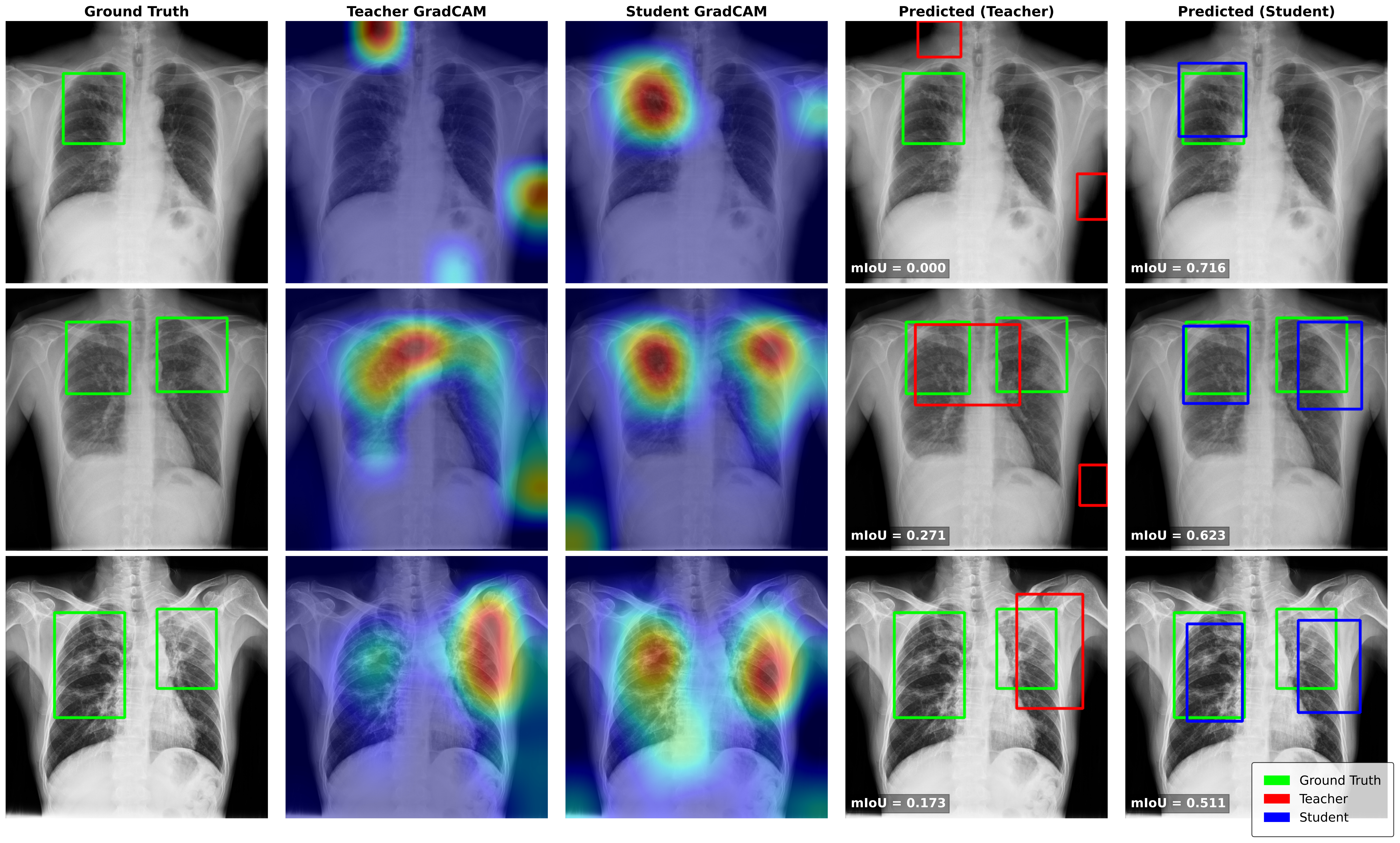}
\caption{Visualization of weakly supervised tuberculosis localization through knowledge distillation. Each row presents a representative test chest X-ray from the TB dataset. Columns (from left to right) show: (1) ground-truth lesion annotations (green), (2) Grad-CAM attention maps from the teacher model, (3) Grad-CAM attention maps from the distilled student model, (4) predicted bounding boxes of the teacher model (red) overlaid with ground truth, and (5) predicted bounding boxes of the student model (blue) overlaid with ground truth. The mean Intersection-over-Union (mIOU) values in the bottom-left corners quantify the spatial overlap between Grad-CAM-derived regions and annotated lesions, demonstrating improved localization performance of the distilled student model under weak supervision.}
\label{fig:merged_gradcam}
\end{figure}

We measure our result both quantitatively and qualitatively. The qualitative result is shown in the Figure \ref{fig:merged_gradcam}. Each row displays an example of a tuberculosis (TB) chest X-ray image taken from the validation subset of the TBX11k dataset. A detailed inspection of these images reveals a significant difference in how the teacher and student models attend to the visual features. Specifically, the teacher model tends to highlight incorrect or irrelevant areas of the chest, suggesting that it is relying on spurious correlations rather than meaningful disease indicators. In contrast, the heatmaps generated by the student model closely align with the ground truth bounding boxes, clearly indicating that it has successfully learned to identify the key features associated with TB lesions. This comparison demonstrates that the student model is better at focusing on diagnostically relevant regions, reflecting a more accurate and clinically useful understanding of TB pathology.\par

\begin{figure}[htbp]
\centering
\includegraphics[width=0.9\linewidth]{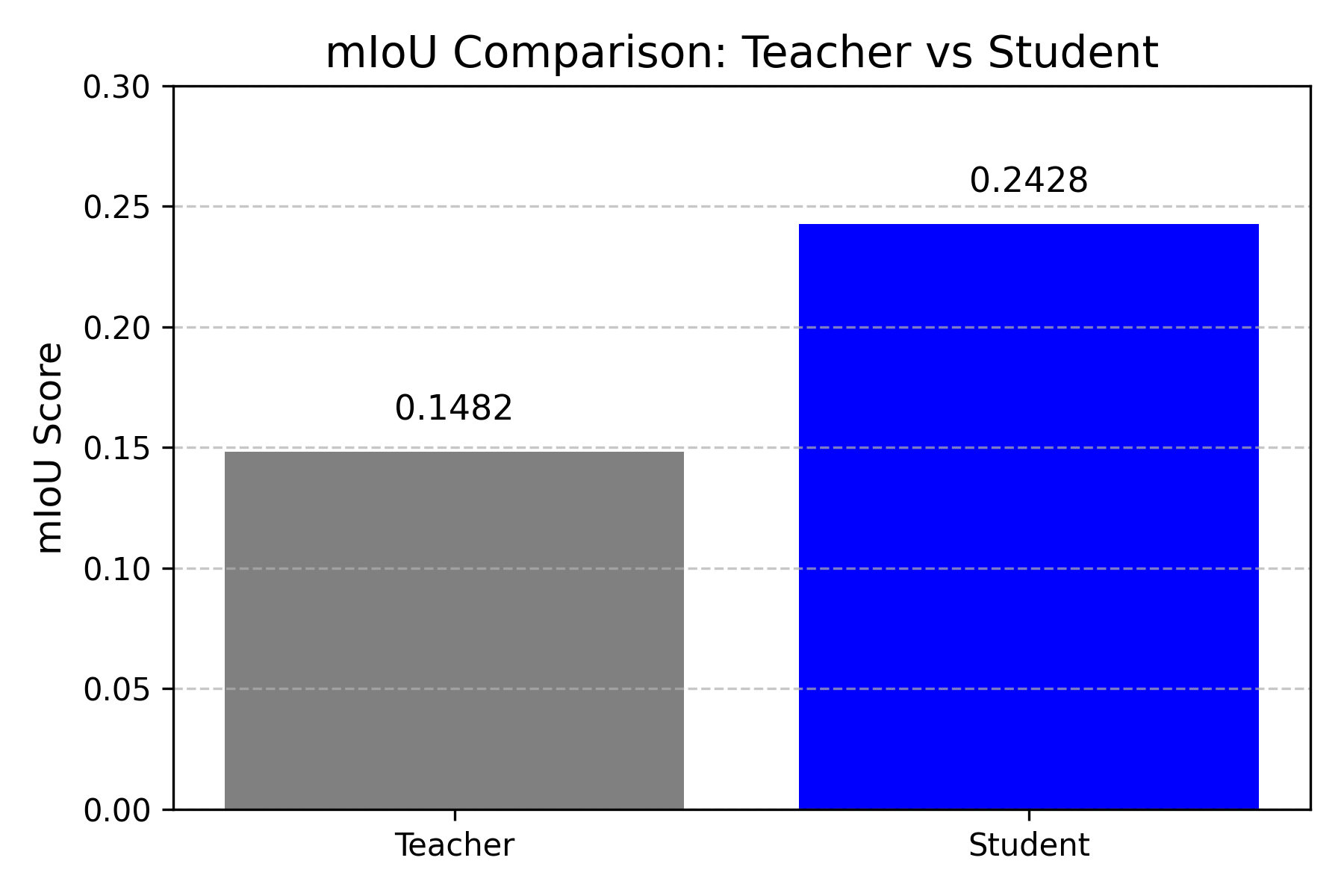}
\caption{The student model achieves a substantially higher mIOU score than the teacher model, indicating that the student is better suited for weakly supervised tuberculosis localization.}
\label{fig:main_comparison}
\end{figure}

We use the metric mIOU (mean intersection over union) devised according to Algorithm 2 and 3 to quantitatively compare the teacher model and the student model. The result is shown in Figure \ref{fig:main_comparison}. \\

\begin{figure}[htbp] 
\centering 
\includegraphics[width=0.45\linewidth]{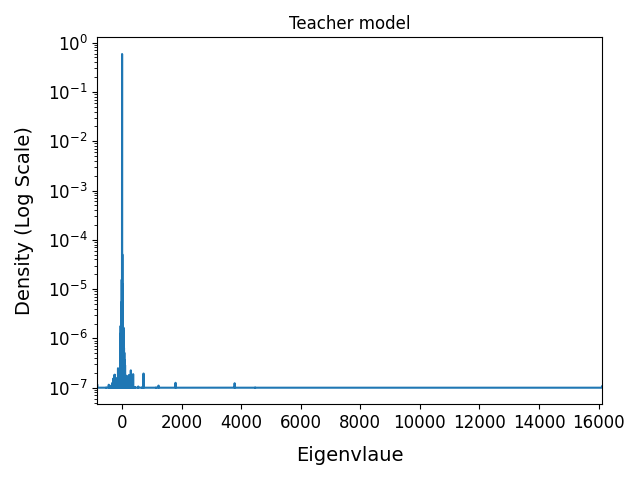} 
\medskip 
\includegraphics[width=0.45\linewidth]{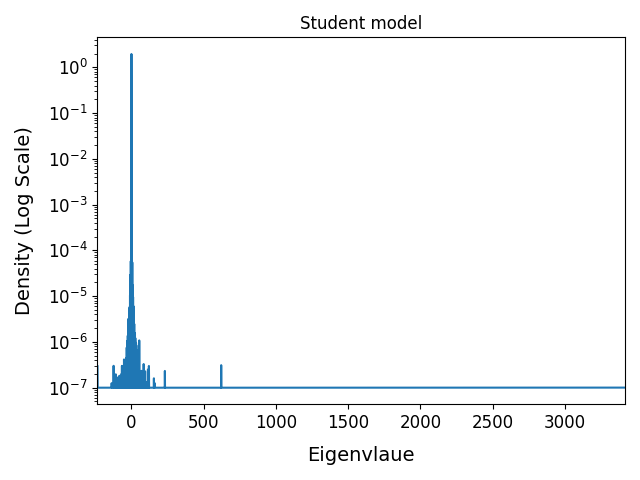} \caption{The ESD plots show that most Hessian eigenvalues are near zero, indicating flat regions in the loss landscape, while a long tail in the teacher model suggests sharper directions.} 
\label{fig:esd_plot} 
\end{figure}

To investigate why knowledge distillation results in higher mIOU, we try to understand the nature of the loss landscape of both the teacher model and the student model via Hessian analysis. The eigenvalues of the Hessian matrix measure the curvature along each direction in the weight space. We computed the Hessian eigenvalues using PyHessian library\cite{Yao2020}. Top 10 eigenvalues are shown in Table \ref{tab:eigenvalues_table} and the trace of the Hessian matrix (sum of eigenvalues) is shown in Table \ref{tab:trace_hessian}. The eigenvalues of the student model and the trace are smaller than that of the teacher model. Flat minima are associated with smaller Hessian eigenvalues. Loss landscape with flat minima generalizes better\cite{Keskar2017}. The flatness of the minima helps the student model to learn TB specific feature (high mIOU score). We also plot the Eigenvalue Spectrum Density (ESD) of both the teacher model and the student model (Figure \ref{fig:esd_plot}). Both ESD plots show that a lot of the eigenvalues of the Hessian are close to zero. This means that a lot of the directions along the loss landscape are almost flat. There are some negative eigenvalues, which indicates that there are a few directions with negative curvature. The long tail in the ESD plot of the teacher model highlights sharper directions. Sharper directions are more susceptible to noise. That is why the teacher model is biased towards spurious correlation (low mIOU score). Together Table \ref{tab:eigenvalues_table} and \ref{tab:trace_hessian} along with Figure \ref{fig:esd_plot} show the effectiveness of knowledge distillation in achieving flatter minima and avoiding spurious correlations.\\

\begin{table}[htbp]
\caption{Top 10 eigenvalues of the Hessian matrix}\label{tab:eigenvalues_table}
\centering
\begin{tabular}{@{}lp{8cm}@{}}
\toprule
\textbf{Teacher} &
16113.1055, 3777.7590, 4470.2427,1793.0648, 1214.0911,\\
& 1148.4277, 843.6339, -813.5349, 716.4574, 658.4104 \\
\midrule
\textbf{Student} &
3414.7578, 620.5816, -234.4829, 230.7972, 158.6752,\\
& 155.0008, 132.6553, -130.6553, 125.0955, -118.5276 \\
\botrule
\end{tabular}
\end{table}

\begin{table}[htbp]
\caption{Trace of the Hessian matrix}\label{tab:trace_hessian}
\centering
\begin{tabular}{@{}lp{5cm}@{}}
\toprule
\textbf{Teacher} & 40864.3358 \\
\midrule
\textbf{Student} & 4064.3874 \\
\botrule
\end{tabular}
\end{table}

\begin{figure}[htbp]
\centering
\includegraphics[width=0.9\linewidth]{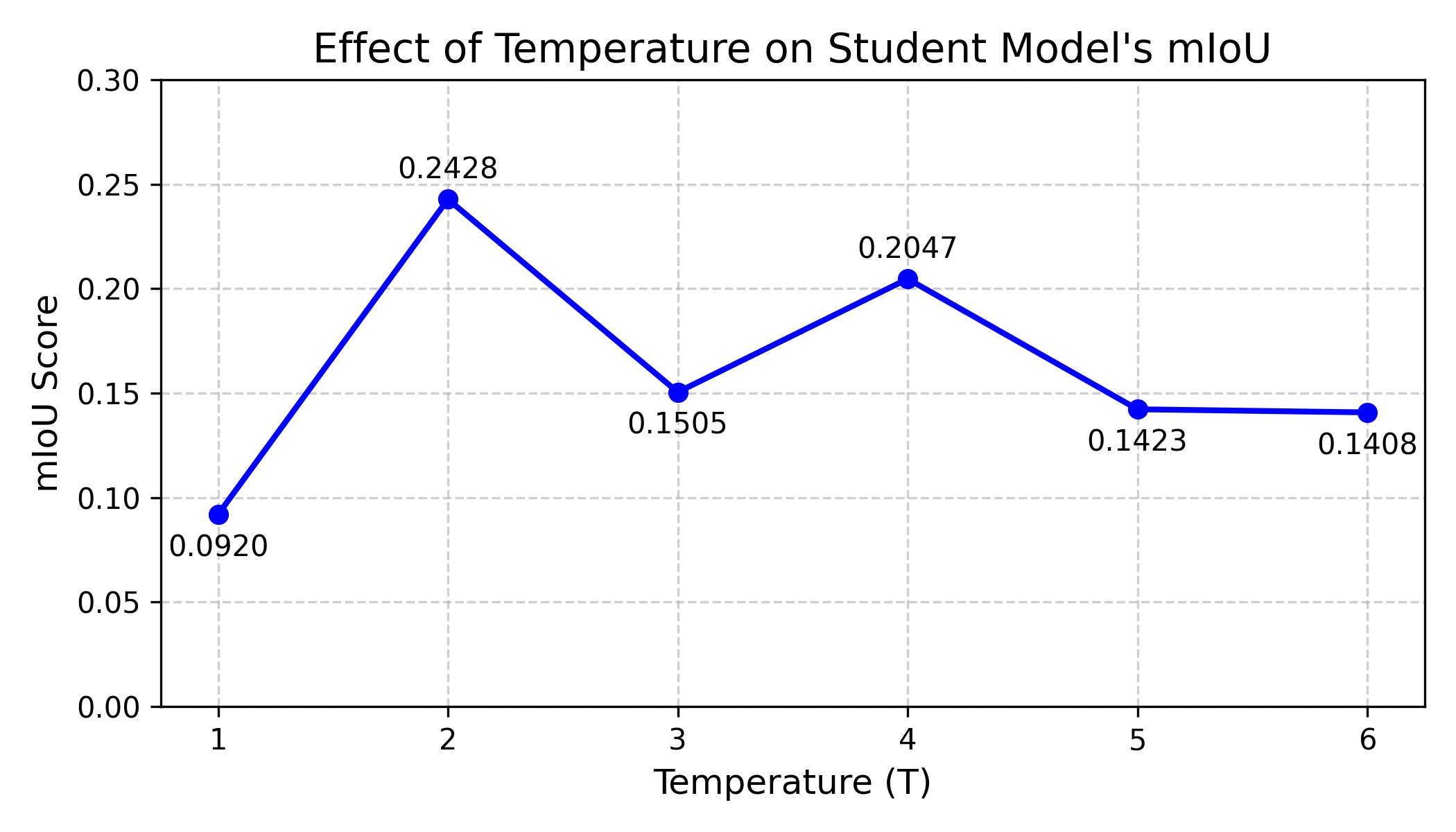}
\caption{The temperature hyperparameter $T$ is systematically varied to evaluate its influence on the student model's mIOU score. Optimal performance is observed at $T = 2$.}
\label{fig:effect_temperature}
\end{figure}

\textit{Effect of temperature (T):} We vary the hyperparameter T to observe its effect on the mIOU score of the student model. The mIOU score peaked for T=2 (Figure \ref{fig:effect_temperature}). The temperature parameter T is used to soften the teacher model's prediction. For T=1, the soft target loss is similar to standard cross entropy loss. For higher values of T (T = 3 or higher),  the distribution becomes too soft and flat. The teacher's output becomes noisy and the student can't distinguish between a confident prediction and a guess. This washes out the signal, hindering learning. At T=2, the teacher's predictions are likely softened just enough to transfer useful information about where to look for TB. This optimal level of softening allows the student model to learn the teacher's intuition effectively. \\

\begin{figure}[htbp]
\centering
\includegraphics[width=0.9\linewidth]{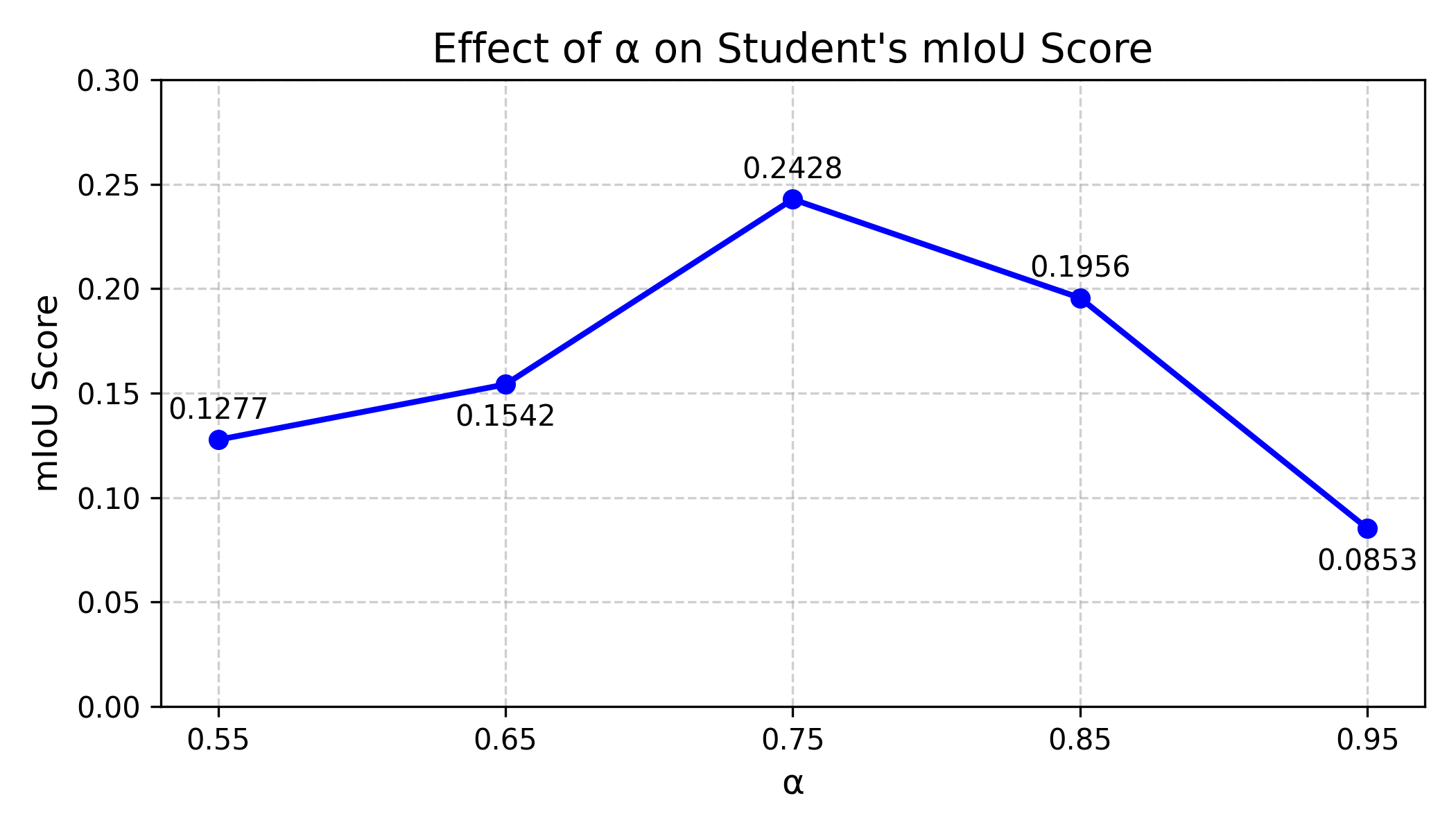}
\caption{The hyperparameter $\alpha$ is systematically varied to evaluate its influence on the student model's mIOU score. Optimal performance is observed at $\alpha = 0.75$.}
\label{fig:effect_alpha}
\end{figure}

\textit{Effect of alpha ($\alpha$):} $\alpha$ is the weightage of cross entropy loss and soft target loss according to Eq. \ref{eq:total_loss}. We systematillcy vary the value of the hyperparameter $\alpha$ from 0.55 to 0.95. We get the highest result for $\alpha = 0.75$ as shown in Figure \ref{fig:effect_alpha}. The teacher model highlights irrelevant areas and relies on spurious correlation. If $\alpha$ is too low, the student will pay too much attention to the flawed teacher and copy its mistakes leading to poor localization and low mIOU. On the other hand, if $\alpha$ is too high the student model will not get enough guidance from the teacher model. $\alpha = 0.75$ works as the best trade-off because it gives the student model enough guidance from the teacher model and also at the same time nudges away from the learned spurious correlation of the teacher model. \\

\textit{Effect of weight initialization: } We discover that mIOU score is heavily tied with the weight initialization of the model. Even the perturbation in only the last layer's weight alters the mIOU score. We fix the dataloader shuffling order and remove other source of randomness to inspect the effect of weight initialization on mIOU which is shown in the Table \ref{tab:mIOU_table}. Nevertheless, the student model beats the teacher model in most cases.\\

\begin{table}[htbp]
\caption{mIOU score comparison between the teacher model and the student model for different seeds}\label{tab:mIOU_table}
\centering
\begin{tabular}{@{}lcc@{}}
\toprule
\textbf{Weight Initialization} & \textbf{Teacher's mIOU} & \textbf{Student's mIOU} \\
\midrule
SEED 0     & 0.1814 & 0.2355 \\
SEED 1     & 0.1324 & 0.1190 \\
SEED 42    & 0.1482 & 0.2428 \\
SEED 1234  & 0.1154 & 0.2651 \\
mean $\pm$ std & 0.1443 $\pm$ 0.0243 & 0.2156 $\pm$ 0.0568 \\
\botrule
\end{tabular}
\end{table}

\textit{Classification result: } Our main goal is to localize the Tuberculosis region in chest x-ray images without bounding box information. Nonetheless, we also ensure that our knowledge distillation method does not harm classification performance. The classification performance of the student model is on par with the teacher model, as can be seen from Table \ref{tab:classification_result}. The classification result shows the need for qualitative comparison between two different models as conventional quantitative comparison fails to capture many nuances across models. Here, the teacher model and the student model achieve similar quantitative performance across different metrics (e.g., accuracy, AUC, sensitivity, specificity, average precision, average recall), though the teacher model relies on spurious correlation for TB prediction, whereas the student model learns robust TB-related features.

\begin{table}[htbp]
\caption{Classification result (TB vs no-TB) comparison}\label{tab:classification_result}
\centering
\begin{tabular}{@{}lcc@{}}
\toprule
\textbf{Metric} & \textbf{Teacher} & \textbf{Student} \\
\midrule
\textbf{Accuracy}          & 0.9817 & \textbf{0.9850} \\
\textbf{AUC}               & 0.9944 & \textbf{0.9954} \\
\textbf{Sensitivity}       & 0.9100 & \textbf{0.9800} \\
\textbf{Specificity}       & \textbf{0.9906} & 0.9856 \\
\textbf{Average precision} & \textbf{0.9563} & 0.9462 \\
\textbf{Average recall}    & 0.9503 & \textbf{0.9828} \\
\botrule
\end{tabular}
\end{table}

These results underscore the first novelty: knowledge distillation's role in reducing spurious correlations, as the student's superior mIOU across seeds indicates refined feature focus. For the second novelty, the generated bounding boxes derived without annotations achieve close overlaps with ground truths. This annotation-free approach could facilitate broader adoption in global health initiatives where expert-labeled data is scarce by enabling rapid model deployment and iterative improvements without recurring annotation expenses.

The model falls into different local minima for different seeds \cite{Swirszcz2017}. Some minima fail to capture the intricate features of the TB CXRs. That is why the mIOU score varies so much for different seeds. However, a more detailed investigation is required and is left for future work.

\section{Conclusion}\label{sec5}
This study demonstrates that using knowledge distillation to train CNN-based models to focus on TB-specific features is both feasible and effective. The student model successfully learned to highlight key TB lesion features, as shown by heatmaps that closely match the ground truth bounding boxes. In contrast, the teacher model often concentrated on irrelevant regions, suggesting it relied on spurious correlations. The reduction of spurious correlations through KD enhances model trustworthiness particularly in heterogeneous clinical environments. More crucially, the ability to generate bounding boxes without annotations addresses a fundamental challenge in medical AI: the high cost of expert labeling. This innovation paves the way for cost-effective, scalable TB diagnostic tools, potentially impacting millions in underserved regions by democratizing access to high-quality datasets and models. The paper also points out that the mean intersection over union (mIOU) score varied notably with different weight initializations (seeds). We recognize that a more thorough investigation into how weight initialization affects the mIOU score is needed and leave this for future work. The main contribution of this study is showing that knowledge distillation can help reduce the impact of spurious correlations in deep learning models, resulting in more reliable and generalizable TB detection.

\bibliography{sn-bibliography}

\end{document}